\documentclass{article}

\usepackage{arxiv}

\usepackage[utf8]{inputenc} 
\usepackage[T1]{fontenc}    
\usepackage{url}            
\usepackage{booktabs}       

\usepackage{amsfonts}       
\usepackage{nicefrac}       
\usepackage{microtype}      
\usepackage{graphicx}
\usepackage{caption}  
\usepackage{subcaption} 

\usepackage[numbers]{natbib}
\usepackage{hyperref}
\usepackage{doi}
\usepackage{booktabs}
\usepackage{multirow}
\usepackage{makecell} 
\usepackage{xcolor}
\usepackage{adjustbox}

\usepackage{tcolorbox}

\title{From Worms to Mice: Homeostasis Maybe All You Need}

\date{December 28, 2024}	
\author{
    \href{https://orcid.org/0000-0001-7914-8494}{\includegraphics[scale=0.06]{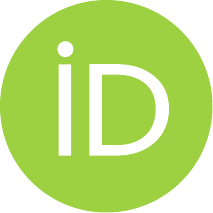}\hspace{1mm}Jesus Marco de Lucas} \\
   \\
    \texttt{jesus.marco@csic.es}
    \\
    Advanced Computing and e-Science Group\\
    Instituto de Física de Cantabria (IFCA) CSIC-Universidad de Cantabria, Santander, ES 39005, SPAIN\\
}

\renewcommand{\headeright}{Short Comment}
\renewcommand{\undertitle}{Short Comment}
\renewcommand{\shorttitle}{\textit{arXiv} From Worms to Mice: Homeostasis Maybe All You Need}

\hypersetup{
pdftitle={From Worms to Mice: Homeostasis Maybe All You Need},
pdfsubject={q-bio.NC, q-bio.QM},
pdfauthor={Jesus.~Marco},
pdfkeywords={XOR, Homeostasis, Connectomes, Neural Motifs, Loss Function },
}

\begin{document}
\maketitle

\begin{abstract}

In this brief and speculative commentary, we explore ideas inspired by neural networks in machine learning, proposing that a simple neural XOR motif, involving both excitatory and inhibitory connections, may provide the basis for a relevant mode of plasticity in neural circuits of living organisms, with homeostasis as the sole guiding principle. This XOR motif simply signals the discrepancy between incoming signals and reference signals, thereby providing a basis for a loss function in learning neural circuits, and at the same time regulating homeostasis by halting the propagation of these incoming signals. The core motif uses a 4:1 ratio of excitatory to inhibitory neurons, and supports broader neural patterns such as the well-known 'winner takes all' (WTA) mechanism. We examined the prevalence of the XOR motif in the published connectomes of various organisms with increasing complexity, and found that it ranges from tens (in C. elegans) to millions (in several Drosophila neuropils) and more than tens of millions (in mouse V1 visual cortex). If validated, our hypothesis identifies two of the three key components in analogy to machine learning models: the architecture and the loss function. And we propose that a relevant type of biological neural plasticity is simply driven by a basic control or regulatory system, which has persisted and adapted despite the increasing complexity of organisms throughout evolution.

\end{abstract}


\section{Introduction}

IHomeostasis is a general term that can be defined as "the ability or tendency of a living organism, cell, or group to maintain the conditions within it at the same level despite changes in the conditions around it, or this state of internal balance"\cite{1}
However, in the context of neuroscience, in addition to conditions at key levels such as intracellular Ca concentration, the term also refers to the equilibrium achieved in the propagation of electrical and chemical activity, preventing unstable conditions such as neural hyperactivity.
In a previous short commentary\cite{2}, we presented a basic neural motif composed of a few excitatory neurons and one inhibitory neuron, corresponding to C. Elegans, and showed that it can implement an XOR switch, i.e. such that when it compares two input signals, its output is non-zero only if these signals are different. We also described how such an XOR motif can be used in a computational neural network as the core of a basic autoencoder, and trained using the so-called Liquid Time Constants (LTC) neural network framework\cite{3}.
In this new note, we extend the previous implementation to the case of spiking neurons, and explore in more detail the presence of this XOR motif in different organisms using the available open data for their connectomes. First, we examine the C. elegans connectome again, but requiring the explicit XOR motif to inhibit connections. Then we look at the recently published connectomes of the different Drosophila neuropils. Finally, we also examine the connectome describing the V1 visual cortex in mice. 
We conclude by discussing the potential relevance of this very simple XOR motif as a core piece for configuring biologically viable computational neural networks.

\section{The XOR motif in spiking neural networks}

\begin{figure}[h!]
	\centering
       \adjustbox{cfbox=black 1pt}{ 
       \includegraphics[width=0.65\textwidth]{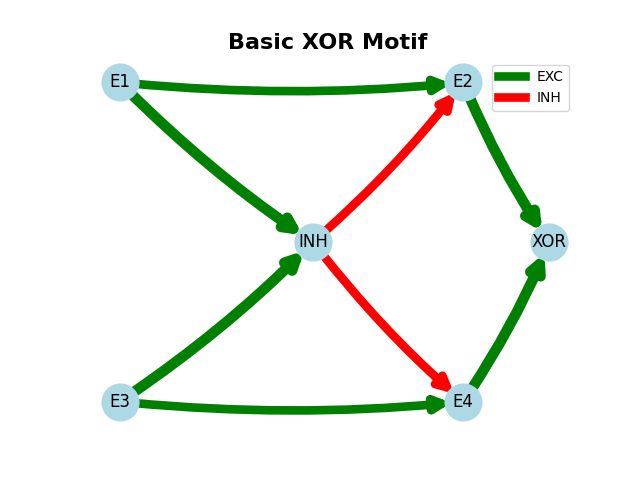} }
	\caption{basic scheme of the neural XOR motif found in C.Elegans.}
	\label{fig:XORmotif}
\end{figure}

We introduced a basic neural motif using C. elegans neurons that implements an XOR switch in a previous note [2]. The scheme is shown in Figure~\ref{fig:XORmotif}: two excitatory neurons, E1 and E3, are each connected to two other excitatory neurons, E2 and E4, but also to an inhibitory interneuron, INH, which is also connected to E2 and E4; these two neurons provide the input to a final neuron, which provides the XOR output.

\begin{figure}[ht!]
    \centering
    \begin{subfigure}{0.495\textwidth}
        \centering
        \includegraphics[width=\textwidth]{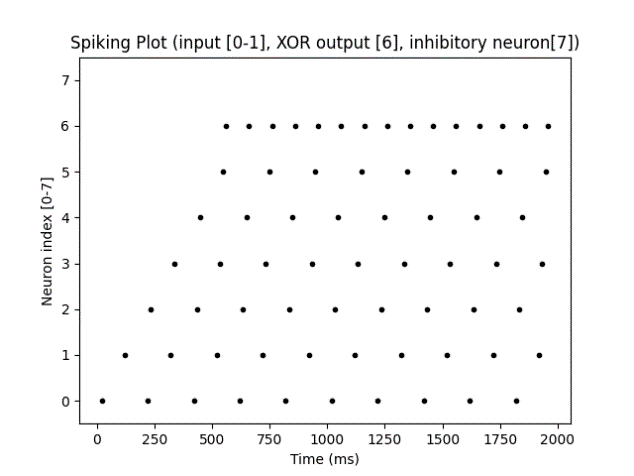}
        \label{fig:2a}
    \end{subfigure}
    \begin{subfigure}{0.495\textwidth}
        \centering
        \includegraphics[width=\textwidth]{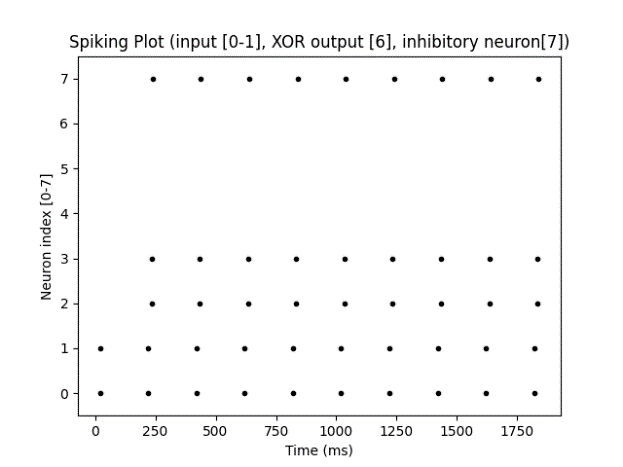}
        \label{fig:2b}
    \end{subfigure}
    \caption{Spiking patterns in the XOR motif neurons (2-7), corresponding to the injection of two pulses (neurons 0 and 1); left: 0-1 / 1-0 pattern, neuron 6 (XOR output) spikes regularly; right: 0-0 / 1-1 pattern, neuron 6 does not spike, as neurons 4 and 5 are inhibited by neuron 7.}
    \label{fig:XOR-Spiking}
\end{figure}

We have implemented a similar scheme but with spiking neurons using the Brian2 platform\cite{4}. To contrast the performance as an XOR switch, we simulated two different pulse trains and used them as input to two sensory neurons coupled to the XOR motif. This XOR motif contains four excitatory neurons, one inhibitory neuron and one output neuron. Figure~\ref{fig:XOR-Spiking}  shows how these pulse trains produce the corresponding spiking pulses in the output neuron following the XOR pattern.

The parameters of this simple simulation are summarized in the box below, and the code is available in github\cite{5}. 

\begin{tcolorbox}[colback=gray!5, colframe=gray!80, boxrule=0.5mm, arc=4mm, sharp corners]
tau\_excit = 10 ms: Excitatory neurons membrane time constant. \\
tau\_inhib = 10 ms: Inhibitory neurons membrane time constant. \\
V\_threshold = -50.0 mV: Firing threshold. \\
V\_rest = -60 mV: Rest Voltage. \\
V\_reset = -65 mV: Restart Voltage after firing. \\
V\_min = -80 mV: Lower limit for membrane voltage (after inhibition). \\
V\_max = +30 mV: Upper limit for membrane voltage. \\
Refractory period: 2 ms. \\
dv/dt = (V\_rest - v + I\_exc - I\_inh) / tau: Voltage dynamics (unless refractory). \\
dI\_exc/dt = -I\_exc / (20 ms): Excitatory current decay. \\
dI\_inh/dt = -I\_inh / (20 ms): Inhibitory current decay.
\end{tcolorbox}

This very simple scheme is implemented only to show the feasibility of an XOR motif for spiking neurons; we reserve the very interesting discussion of time-dependent plasticity, associativity and inference for a new note\cite{6}.

\section{XOR Motif in C. Elegans}

In this section, we review and extend previous findings regarding the presence and relevance of the XOR motif in the connectome of C. Elegans\cite{2}.
We utilized reference data from the adult hermaphrodite connectome\cite{7}, available as an adjacency matrix in the WormWiring platform\cite{8}, which includes 3.707 directed chemical synaptic connections among neurons. From this dataset, we compiled a table of the 300 neurons involved in these connections and assigned a putative role (excitatory, inhibitory, or other) to each neuron based on its associated neurotransmitter. The neurotransmitter data was derived from the WormAtlas platform\cite{9} and additional sources\cite{10}.
Neurons classified as “excitatory” include glutamatergic and cholinergic neurons, while “inhibitory” neurons are primarily GABAergic. Neurons using serotonin, octopamine, dopamine, or orphan neurons without a known neurotransmitter were categorized as "other."
Using this updated neuron classification, we confirmed the presence of the XOR motif in the adult hermaphrodite connectome and extended our previous results by explicitly requiring specific neuron types in the different nodes of the motif.

\subsection{Identifying XOR Motifs in the C. Elegans Connectome}
Searching for a small subgraph motif, like the XOR motif (a six-node configuration, each node potentially of three types, with eight directed edges), within a larger graph such as the C. Elegans connectome (300 nodes and over 3.700 edges), is facilitated by graph analysis tools such as the NetworkX Python library\cite{11}, particularly its directed graph (DiGraph) methods.
First, as in our previous work, we identified the “strict” XOR motif, defined as six neurons connected exclusively through eight specific edges, without imposing additional constraints on neuron types. Using the isomorphism matching functionality in DiGraph and removing duplicates from symmetric configurations 
(e.g. interchanging nodes 1$\leftrightarrow$3 and 2$\leftrightarrow$4 ), we found 722 strict XOR motifs. This count is significantly higher than the 285 motifs reported in our earlier analysis, which used a connectome with only 2.100 edges.
Next, we imposed neuron-type constraints on the XOR motif nodes: node 6 as inhibitory, nodes 1–4 as excitatory, and no restriction on node 5. To enhance analytical coverage, we also allowed neurons of the "other" type to occupy any node. With these restrictions, we identified 134 "true" XOR motifs, confirming the significance of this motif under biologically relevant constraints, particularly as it corresponds to a configuration capable of implementing a functional XOR switch.

\subsection{Virtual XOR Motifs in the Connectome}
The NetworkX matcher.subgraph.isomorphism iterator has proved invaluable for identifying motifs in large graphs, and we have used it extensively in the analysis of other connectomes, as discussed below.
However, for a relatively small connectome, such as that of C.Elegans, we also have the possibility of testing "virtual" XOR motifs, i.e. motifs involving additional edges between the same nodes. Our main interest is to extend the "true" XOR motif with connections providing feedback from the output neuron, to provide a basic plasticity mechanism by reinforcement through this feedback, as we already suggested in our previous note\cite{2} and will discuss below. We used another Digraph procedure, the matcher.subgraph.monomorphism iterator, to find these "virtual" motifs.

\begin{figure}[ht!]
    \centering
    \begin{subfigure}{0.49\textwidth}
        \centering
        \includegraphics[width=\textwidth]{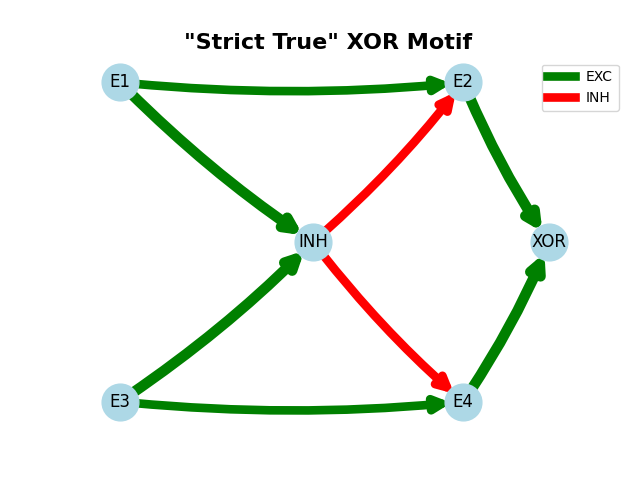}
        \label{fig:3a}
    \end{subfigure}
    \begin{subfigure}{0.49\textwidth}
        \centering
        \includegraphics[width=\textwidth]{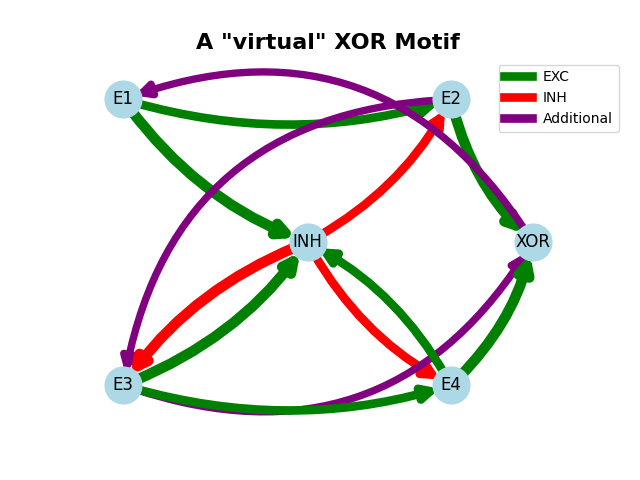}
        \label{fig:3b}
    \end{subfigure}
    \caption{Comparison of a basic “strict” XOR motif (left) and an example of “virtual” one (right), that includes additional connections among the nodes. Strict motifs are defined as an isomorphism between the motif and a subgraph in the connectome, while virtual motifs are defined as monomorphisms.}
    \label{fig:StrictExtended-XOR}
\end{figure}

Figure~\ref{fig:StrictExtended-XOR} shows graphically the difference between “strict” and “virtual” XOR motifs.

\begin{figure}[h!]
	\centering
       \adjustbox{cfbox=black 1pt}{ 
       \includegraphics[width=0.65\textwidth]{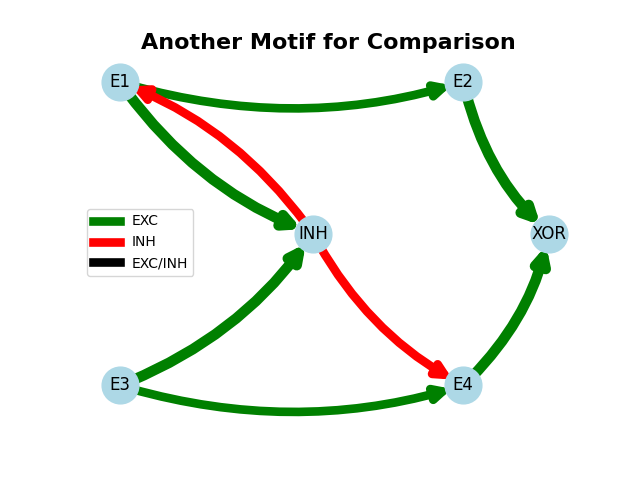} }
	\caption{A different motif with six interconnected neurons, one of them inhibitory, and eight edges. Notice that other configurations may overlap with the strict XOR motif, so it is not obvious what other motifs could be included in the comparison.}
	\label{fig:different-XOR}
\end{figure}

There is a very large number of these "virtual" XOR motifs in the connectome, more than 2 million, if no constraint on the type of node is imposed, and even if we impose the restrictive configuration of a "true" XOR motif, i.e. that neurons E1-E4 are excitatory and neuron INH is inhibitory, we find quite a large number of these "virtual true" XOR motifs in the connectome: 82.558. Somewhat surprisingly, most of them correspond to only five GABA inhibitory interneurons: AVJR (involved in 52,555 motifs), AVJL (19,398), RIBR/L (5,507/3,358) and RIS (970). 
It should be noted that these "virtual" XOR motifs can only be considered as such if the external inputs arrive via the two sensory neurons in the motif (labelled E1, E3 in the figure). In less than 3\% (2,399 cases) of these "virtual" motifs, the XOR output neuron is also inhibitory (AVJL/R and AVL in 70\% of cases), while the 97\% of excitatory XOR neurons are mainly interneurons (AVA L/R, AVE L/R, AVD L/R, AVB L/R) and motoneurons (DA1-9, DB2-7, AS1-11), as expected for a feedforward network.
For comparison, we also tested the number of other similar motifs, including a single inhibitory neuron, but with a different arrangement of its connections (see Figure~\ref{fig:different-XOR} ), and found 49,174 such motifs, significantly fewer.

\begin{figure}[ht!]
    \centering
    \begin{subfigure}{0.49\textwidth}
        \centering
        \includegraphics[width=\textwidth]{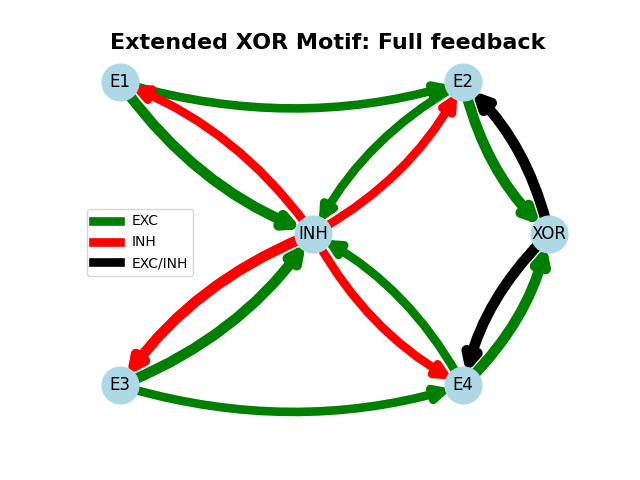}
        \label{fig:FullFeedback-XOR}
    \end{subfigure}
    \begin{subfigure}{0.49\textwidth}
        \centering
        \includegraphics[width=\textwidth]{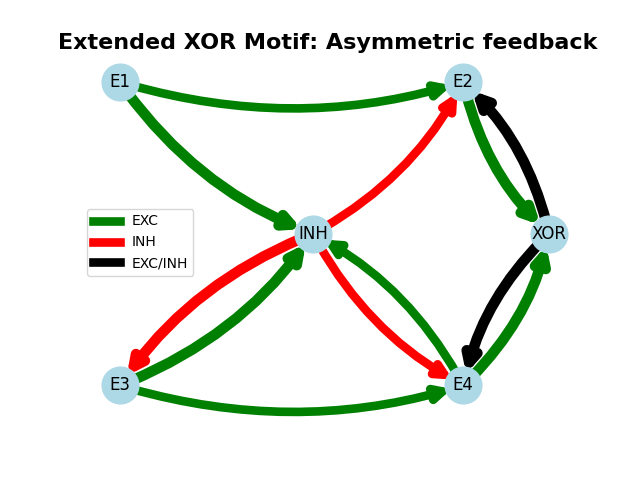}
        \label{fig:AsymFeedback-XOR}
    \end{subfigure}
    \caption{Two “extended” XOR motifs: left, full feedback; right, asymmetric feedback..}
    \label{fig:FullAsym-XOR}
\end{figure}

We can now begin to explore 'extended' XOR motifs, where feedback is provided from the output of the XOR neuron to other neurons, which are in fact a subset of the 'virtual' XOR motifs. We consider the configurations where the feedback path connects this XOR neuron back to the two excitatory neurons connected to it (i.e. those labelled 2 and 4), and these in turn connect back to the inhibitory interneuron, which also connects back to the input excitatory neurons, as shown in Figure~\ref{fig:FullAsym-XOR} .

The virtual configurations corresponding to this “extended” XOR full feedback motif appear 279 times in the C.Elegans connectome, a number that increases to 2.001 times if we also consider neurons of the "other" type in any of the nodes.
Finally, following the ideas on the implementation of an autoencoder core presented in our previous note, we also considered an asymmetric configuration, where there is no feedback connection from E2 to INH, nor from INH to E1 (see Figure~\ref{fig:FullAsym-XOR} ): there are 4.425 virtual implementations of such an extended XOR motif in the C. Elegans adult hermaphrodite connectome, a number that increases to 16.917 when including "other" neurons.  

Although it is not possible to clearly infer a privileged recurrent network configuration in such a dense configuration with so many embedded virtual XOR motifs, these results confirm the interest of further research on these configurations, using the XOR output to provide feedback to the whole motif connections.

We reserve for a future note the discussion of the relevance of embodiment, as suggested in\cite{12}, i.e. the patterns that connect these XOR motifs to the sensory and motor neurons.

\section{XOR motif in Drosophila}

The recent publication of the complete connectome of Drosophila\cite{13}, has allowed us to extend the analysis of XOR motif abundance to a much more complex organism with very different neurons. We have used the data corresponding to 80 different neuropils available from\cite{14}, and have considered cholinergic connections as excitatory and both GABAergic and glutamatergic connections as inhibitory, following \cite{15},\cite{16},\cite{17}.

\begin{figure}[h!]
	\centering
       \adjustbox{cfbox=black 1pt}{ 
       \includegraphics[width=0.8\textwidth]{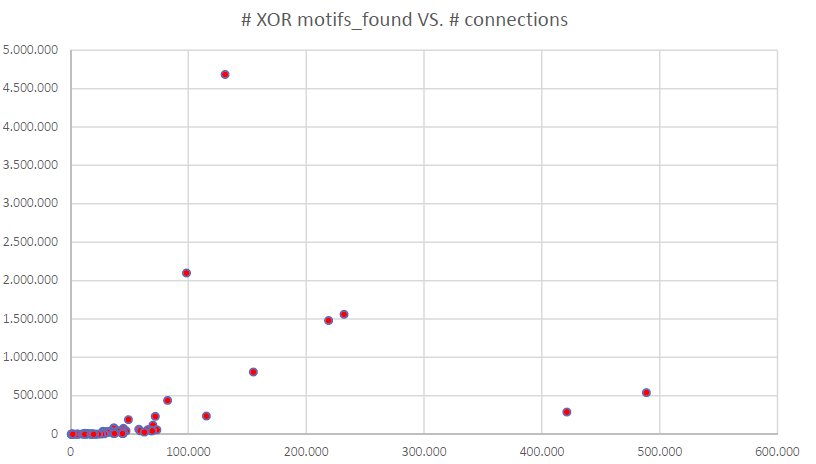} }
	\caption{Number of “true strict” XOR motifs found in different neuropils versus number of connections established between neurons in those neuropils.}
	\label{fig:RatioDrosophila}
\end{figure}

Using a parallelized Python script and restricting our search to isomorphisms, we examined the number of "strictly true" XOR motifs in the 80 neuropils. The results are presented in Table~\ref{tab:neuropil_data} below for the 30 neuropils with more neurons. It can be observed that the ratio of XOR motifs increases significantly for some of the neuropils, and it is strictly zero in other cases, as also shown in Figure~\ref{fig:RatioDrosophila} .

\begin{table}[ht]
\centering
\begin{tabular}{lrrrr}
\toprule
\textbf{NEUROPIL} & \textbf{\# neurons} & \textbf{\# conn.} & \textbf{\# XOR} & \textbf{motifs} \\
\midrule
ME\_R   & 36.142 & 488.839 & 540.676    & \\
ME\_L   & 35.624 & 421.151 & 288.024    & \\
LO\_L   & 22.364 & 219.046 & 1.480.722  & \\
LO\_R   & 22.057 & 232.176 & 1.560.212  & \\
LOP\_R  & 11.796 & 115.374 & 234.626    & \\
LOP\_L  & 11.211 & 68.368  & 43.788     & \\
GNG     & 10.088 & 155.111 & 808.052    & \\
LA\_R   & 8.736  & 17.543  & 0          & \\
PLP\_L  & 7.823  & 73.308  & 54.912     & \\
PLP\_R  & 7.383  & 65.621  & 57.840     & \\
SAD     & 6.775  & 68.539  & 44.812     & \\
LA\_L   & 6.570  & 10.482  & 0          & \\
SMP\_R  & 6.294  & 69.331  & 42.284     & \\
SMP\_L  & 6.217  & 62.718  & 25.412     & \\
SLP\_R  & 5.988  & 59.171  & 41.526     & \\
SLP\_L  & 5.753  & 44.015  & 8.674      & \\
AVLP\_R & 5.703  & 131.152 & 4.683.376  & \\
SCL\_R  & 5.542  & 44.354  & 37.888     & \\
SPS\_R  & 5.538  & 69.996  & 117.470    & \\
PVLP\_L & 5.446  & 82.315  & 438.416    & \\
CRE\_L  & 5.196  & 36.450  & 7.914      & \\
CRE\_R  & 5.067  & 37.359  & 5.884      & \\
SCL\_L  & 4.955  & 37.251  & 17.582     & \\
SPS\_L  & 4.917  & 57.941  & 62.318     & \\
PVLP\_R & 4.836  & 71.855  & 231.460    & \\
ICL\_R  & 4.567  & 44.900  & 74.374     & \\
AVLP\_L & 4.493  & 98.418  & 2.098.224  & \\
SIP\_L  & 4.399  & 26.299  & 2.246      & \\
ICL\_L  & 4.178  & 40.552  & 44.598     & \\
\bottomrule
\end{tabular}
\caption{Statistics of “true strict” XOR motifs found in different neuropils in the Drosophila connectome.}
\label{tab:neuropil_data}
\end{table}

It is interesting that the neuropils in which the ratio of XOR motifs is higher compared to the number of neuronal connections are involved in the processing of sensory input: the AVLP (Anterior Ventrolateral Protocerebrum) L/R neuropils are an integrative centre for sensory and motor information, whereas the LO (Lamina Optic) L/R are involved in the processing of visual information. It is also interesting to note the clear asymmetry between AVLP\_R and AVLP\_L, while the number of neurons and connections are quite similar.

\section{XOR motifs in mice}

To continue the study of connectomes in other more complex organisms, we used the open data available from the Brain Allen Institute \cite{18} to analyse the V1 visual cortex of mice, which contains a total of around 79 million connections between 230.924 neurons (192.545 excitatory neurons, 38.346 inhibitory neurons), and was used to configure and simulate a network of GLIF (Leak Integrate and Fire) neurons, which was compared with another implemented using a more detailed and realistic biological model of the neurons\cite{19}. 
The input data is structured as an h5 file, where a subset describing the connectome edges refers to another one assigning all neurons an identification and type, and a third structure describes a total of 111 neuron types, covering the L2/3, L4, L5 and L6 V1 layers. We have transformed this h5 file structure into three .csv files ( \texttt{v1\_edges.csv}
, \texttt{v1\_nodes.csv},\texttt{v1\_nodes\_types.csv} ) so we can reuse the previous python scripts with minor modifications. 
However, unlike the previous case, where we have divided the drosophila connectome into neuropils, there is no obvious segmentation that we could use to explore this quite large connectome, nor an obvious way to parallelize such exploration.
Exploring the “first” 10 million connections, following the ordering in the original list of edges, including 230.893 neurons, we have found 34.524.437 isomorphic instances of the “true strict” XOR motif. This is not only an incomplete but also a biased result, as it is affected by such ordering and the potential connection to a specific substructure, and in fact when using another subset of 10 million of connections randomly selected among the 79 million in the total sample, we have found that the number of “true strict” XOR motifs was decreased by almost a factor 5 (7.034.330 isomorphic instances).
As processing the full connectome may require several months with our current approach, we have started to explore the layered structure for these sets of “true strict” XOR motifs. Neurons’ names in this V1 set are assigned according to their excitatory or inhibitory nature (e/i) and their cortex layers (2/3, 4, 5, 6). Tables~\ref{tab:neuron_motifs} and~\ref{tab:neuron_motifs_2} show the neuron types for each node (E1/E3, E2/E4, INH and XOR) in the XOR motifs found in the two subsets of 10 million connections. 

\begin{table}[ht]
\centering
\resizebox{\textwidth}{!}{%
\begin{tabular}{lrlrlrlrl}
\toprule
\textbf{E1/E3 neuron} & \textbf{\# in motifs} & 
\textbf{E2/E4 neuron} & \textbf{\# in motifs} & 
\textbf{INH neuron} & \textbf{\# in motifs} & 
\textbf{XOR neuron} & \textbf{\# in motifs} \\
\midrule
e23Cux2    & 11.459.611 & e5Rbp4     & 33.565.657 & i5Pvalb    & 34.524.436 & e6Ntsr1    & 26.679.839 \\
e5Rbp4     & 7.963.210  & e6Ntsr1    & 610.311    &            &            & e5Rbp4     & 7.119.814  \\
e4other    & 4.415.809  & e5noRbp4   & 348.468    &            &            & e5noRbp4   & 511.857    \\
e4Scnn1a   & 4.059.829  &            &            &            &            & i5Pvalb    & 145.217    \\
e4Rorb     & 3.655.015  &            &            &            &            & e4Scnn1a   & 46.589     \\
e4Nr5a1    & 1.563.906  &            &            &            &            & e4other    & 18.328     \\
e5noRbp4   & 1.330.380  &            &            &            &            & e4Nr5a1    & 2.792      \\
e6Ntsr1    & 76.676     &            &            &            &            &            &            \\
\bottomrule
\end{tabular}%
}
\caption{Identification of the main types of neurons in each node in the XOR motif in the first (original order) subset of 10M of connections.}
\label{tab:neuron_motifs}
\end{table}

\begin{table}[ht]
\centering
\resizebox{\textwidth}{!}{%
\begin{tabular}{lrlrlrlrl}
\toprule
\textbf{E1/E3 neuron} & \textbf{\# in motifs} & 
\textbf{E2/E4 neuron} & \textbf{\# in motifs} & 
\textbf{INH neuron} & \textbf{\# in motifs} & 
\textbf{XOR neuron} & \textbf{\# in motifs} \\
\midrule
e4Scnn1a   & 3.610.439  & e23Cux2    & 4.603.224  & i4Pvalb    & 3.063.020  & e23Cux2    & 2.822.879 \\
e4other    & 3.838.831  & e4Scnn1a   & 2.803.866  & i4Sst      & 2.114.102  & e5Rbp4     & 620.010   \\
e4Rorb     & 3.134.477  & e4Rorb     & 2.409.136  & i4Htr3a    & 1.027.815  & e4other    & 564.261   \\
e23Cux2    & 1.966.040  & e4Nr5a1    & 1.098.249  & i23Pvalb   & 603.708    & e4Scnn1a   & 530.667   \\
e4Nr5a1    & 1.428.095  & e4other    & 2.976.817  & i23Sst     & 156.851    & e4Rorb     & 456.520   \\
e5Rbp4     & 72.126     & e5Rbp4     & 142.907    & i5Sst      & 46.263     & i4Sst      & 443.544   \\
e5noRbp4   & 18.305     & e5noRbp4   & 34.136     & i23Htr3a   & 14.767     & i4Pvalb    & 397.148   \\
e6Ntsr1    & 347        & e6Ntsr1    & 325        & i5Pvalb    & 7.600      & i23Pvalb   & 260.466   \\
           &            &            &            & i5Htr3a    & 88         & i4Htr3a    & 225.908   \\
           &            &            &            & i6Pvalb    & 82         & e4Nr5a1    & 213.399   \\
           &            &            &            & i1Htr3a    & 19         & e5noRbp4   & 154.917   \\
           &            &            &            & i6Sst      & 11         & i23Htr3a   & 100.174   \\
           &            &            &            & i6Htr3a    & 4          & i23Sst     & 76.445    \\
           &            &            &            &            &            & i5Sst      & 69.890    \\
           &            &            &            &            &            & e6Ntsr1    & 53.869    \\
           &            &            &            &            &            & i5Pvalb    & 42.934    \\
           &            &            &            &            &            & i5Htr3a    & 1.052     \\
           &            &            &            &            &            & i6Sst      & 114       \\
           &            &            &            &            &            & i6Pvalb    & 94        \\
           &            &            &            &            &            & i1Htr3a    & 24        \\
           &            &            &            &            &            & i6Htr3a    & 15        \\
\bottomrule
\end{tabular}%
}
\caption{Identification of the main types of neurons in each node in the XOR motif in a subset of 10M of connections randomly shuffled from the whole connectome.}
\label{tab:neuron_motifs_2}
\end{table}

The first set, including the original first 10M connections, surprisingly shows that all 34M motifs are established around only one kind of inhibitory neuron, i5Pvalb (Parvalbumin in layer 5), although there are only 3.746 neurons of these kind in the whole V1 connectome. Let’s remember that PV neurons are fast spiking GABAergic interneurons, expected to help maintaining cortical balance, and ensuring precise temporal coordination in visual processing.  It also shows the relevance of the motifs with an output XOR neuron in layer 6, while the input E1/E3 neurons are much more distributed across all layers, but the intermediate neurons, E2/E4, are mainly in L5.

On the other hand, considering another 10M sample built by a shuffling procedure on the 79M of connections, provides a more diverse structure of XOR motifs, with inhibitory interneurons from all layers, and in particular a majority from L4, including the three main type of INH interneurons: PV, i.e. Pvalb (Parvalbumin), SOM, i.e. Sst (Somatostin), and those that have Htr3a as the defining marker, but may also express other markers like VIP (Vasoactive Intestinal Peptide). 

Confirming these findings, we have also explored in detail the layered structure of these XOR motifs, that is completely different between these two connectome subsets analysed, as shown in tables~\ref{tab:layer_motifs} and~\ref{tab:layer_motifs_2} 

\begin{table}[!htbp]
\centering
\begin{tabular}{lllllll}
\toprule
\textbf{L(E1)} & \textbf{L(E2)} & \textbf{L(E3)} & \textbf{L(E4)} & \textbf{L(XOR)} & \textbf{L(INH)} & \textbf{\# motifs} \\
\midrule
L4   & L5   & L4   & L5   & L6   & L5   & 3.617.382 \\
L4   & L5   & L23  & L5   & L6   & L5   & 3.158.068 \\
L23  & L5   & L4   & L5   & L6   & L5   & 3.148.910 \\
L23  & L5   & L23  & L5   & L6   & L5   & 2.603.375 \\
L5   & L5   & L4   & L5   & L6   & L5   & 2.447.970 \\
L5   & L5   & L23  & L5   & L6   & L5   & 2.025.678 \\
L4   & L5   & L5   & L5   & L6   & L5   & 1.869.381 \\
L23  & L5   & L5   & L5   & L6   & L5   & 1.536.692 \\
L4   & L5   & L4   & L5   & L5   & L5   & 1.286.444 \\
L5   & L5   & L5   & L5   & L6   & L5   & 1.215.356 \\
L4   & L5   & L23  & L5   & L5   & L5   & 1.110.598 \\
L23  & L5   & L4   & L5   & L5   & L5   & 1.104.495 \\
L4   & L5   & L4   & L6   & L6   & L5   & 957.894   \\
L23  & L5   & L23  & L5   & L5   & L5   & 905.141   \\
L5   & L5   & L4   & L5   & L5   & L5   & 863.867   \\
L23  & L5   & L4   & L6   & L6   & L5   & 861.487   \\
L4   & L5   & L5   & L6   & L6   & L5   & 710.778   \\
L5   & L5   & L23  & L5   & L5   & L5   & 706.891   \\
L4   & L5   & L5   & L5   & L5   & L5   & 671.485   \\
L23  & L5   & L5   & L6   & L6   & L5   & 592.040   \\
L5   & L5   & L4   & L6   & L6   & L5   & 578.953   \\
L23  & L5   & L5   & L5   & L5   & L5   & 547.379   \\
L5   & L5   & L5   & L5   & L5   & L5   & 433.509   \\
L5   & L5   & L5   & L6   & L6   & L5   & 427.319   \\
L5   & L6   & L4   & L5   & L6   & L5   & 155.083   \\
L5   & L6   & L23  & L5   & L6   & L5   & 127.663   \\
L4   & L5   & L6   & L6   & L6   & L5   & 101.763   \\
\bottomrule
\end{tabular}
\caption{Identification of the layered structure of the most abundant XOR motif in the first (original order) subset of 10M of connections.}
\label{tab:layer_motifs}
\end{table}

Table~\ref{tab:layer_motifs} confirms the structure already hinted in table~\ref{tab:neuron_motifs}: there are many motifs with the INH neuron in layer 5 and a XOR output neuron in layer 6, with input neurons (E1/E3) from layers L2/3 and L4, and intermediate neurons (E2/E4) in L5.  It is interesting to notice that the pattern in these motifs is quite clear: there are no XOR outputs in L2/3 and very few in L4.

\begin{table}[!htbp]
\centering
\begin{tabular}{lllllll}
\toprule
\textbf{L(E1)} & \textbf{L(E2)} & \textbf{L(E3)} & \textbf{L(E4)} & \textbf{L(XOR)} & \textbf{L(INH)} & \textbf{\# motifs} \\
\midrule
L4   & L4   & L4   & L4   & L4   & L4   & 2.352.156 \\
L4   & L23  & L4   & L4   & L23  & L4   & 989.754   \\
L4   & L4   & L4   & L4   & L23  & L4   & 795.989   \\
L23  & L23  & L23  & L23  & L23  & L23  & 357.301   \\
L4   & L23  & L4   & L23  & L23  & L4   & 324.488   \\
L4   & L4   & L4   & L4   & L5   & L4   & 293.459   \\
L4   & L23  & L4   & L4   & L4   & L4   & 251.163   \\
L4   & L23  & L4   & L4   & L5   & L4   & 244.469   \\
L23  & L23  & L4   & L4   & L23  & L4   & 240.719   \\
L23  & L23  & L4   & L23  & L23  & L23  & 115.022   \\
L23  & L23  & L4   & L23  & L23  & L4   & 96.499    \\
L23  & L4   & L4   & L4   & L4   & L4   & 72.248    \\
L23  & L23  & L4   & L4   & L4   & L4   & 64.122    \\
L23  & L23  & L4   & L4   & L5   & L4   & 60.745    \\
L23  & L23  & L23  & L23  & L5   & L23  & 60.062    \\
L4   & L23  & L23  & L23  & L23  & L23  & 59.202    \\
L4   & L23  & L23  & L23  & L23  & L4   & 50.469    \\
L4   & L23  & L4   & L23  & L5   & L4   & 49.139    \\
L4   & L4   & L4   & L4   & L6   & L4   & 41.301    \\
L23  & L23  & L4   & L4   & L23  & L23  & 26.005    \\
L4   & L5   & L4   & L4   & L5   & L4   & 24.539    \\
L23  & L4   & L4   & L4   & L23  & L4   & 23.549    \\
L4   & L23  & L4   & L23  & L23  & L23  & 22.750    \\
L23  & L23  & L4   & L23  & L5   & L23  & 18.798    \\
L23  & L23  & L23  & L23  & L23  & L4   & 16.229    \\
L23  & L23  & L4   & L23  & L5   & L4   & 14.876    \\
L5   & L23  & L4   & L4   & L23  & L4   & 12.463    \\
L23  & L23  & L23  & L23  & L4   & L23  & 11.662    \\
L4   & L5   & L4   & L4   & L23  & L4   & 11.128    \\
L5   & L4   & L4   & L4   & L4   & L4   & 10.501    \\
L4   & L5   & L4   & L4   & L4   & L4   & 10.060    \\
\bottomrule
\end{tabular}
\caption{Identification of the layered structure of the XOR motif in a subset of 10M of connections randomly shuffled from the whole connectome.}
\label{tab:layer_motifs_2}
\end{table}

As also expected from the previous tables~\ref{tab:neuron_motifs} and~\ref{tab:neuron_motifs_2}, in the second connectome subset we find a much more diverse layered structure, showing the relevance of intralayer motifs in L4 and L2/3.

If we compare this structure with a recent analysis at the microcircuit model in the somatosensory cortex\cite{20}, we may expect to see a more complex excitatory/inhibitory pattern than the simple “true” XOR motifs, where the inhibitory core includes the three types of inhibitory neurons, PV, SOM and VIP.

\section{Discussion}

The analysis of the presence of XOR motifs in very different connectomes, from worms to mice, indicates a potential interest of this explicit configuration involving excitatory and inhibitory neurons, already suggested by previous results\cite{22} considering three and four node motifs in different connectomes\cite{13},\cite{24},\cite{25} and also by the importance of the so-called WTALL (Winner Takes All) circuits \cite{26}.

However, the analysis presented here proposes that the XOR motif may provide a common basis for better understanding the importance of the E/I balance for homeostasis, and probably its relevance as a direct feedback signal for learning, implementing a kind of basic loss function.

Although the relevance of computation at the dendritic level is well established \cite{27},\cite{28},\cite{29}, these XOR motifs may provide the key to computation at the neuronal circuit level. This may be the case in the mouse primary visual cortex, where a GLIF network based on the same connectome as we studied provides firing rates comparable to those obtained from biologically detailed models using NEST\cite{19}. The integration of inhibitory and excitatory neurons in the motif suggests a different approach to solving the XOR problem than other proposals using spiking neurons\cite{30}.

The possibility that a simple regulatory control can also drive a quite sophisticated task such as learning has yet to be confirmed by understanding how such XOR feedback is reinjected and used to promote plasticity, but it seems computationally feasible based on the well-known results of using random feedback to train computational neural networks \cite{31}. 

If so, we could already have two of the three key components proposed in\cite{32} to provide a connecting framework with the machine learning techniques: architecture, and objective function. For the third one, learning methods, we are already exploring a possible implementation in spiking neural networks through backpropagation in time via axonal delays\cite{33}.

\end{document}